\documentclass[11pt, a4paper]{article}

\usepackage[whole]{bxcjkjatype}

\usepackage{bm, amsmath, amsthm, amssymb, accents, comment}
\usepackage{ascmac}
\RequirePackage{amsthm,amsmath,amsfonts,amssymb}
\RequirePackage{natbib,url}
\RequirePackage{graphicx}
\usepackage{tikz}
\usepackage{pgfplots}
\usepackage{subcaption}
\pgfplotsset{
	compat=newest, 
	cycle list name=exotic }
\usepackage{algorithm}
\usepackage{algpseudocode, setspace}

\newcommand{\argmax}{\mathop{\rm argmax}\limits}

\usepackage{typearea}
\typearea{12}

\begin{document}
	
	\title{Empirical Bayes 1-bit matrix completion}
	
	\author{Takeru Matsuda\thanks{Department of Mathematical Informatics, Graduate School of Information Science and Technology, The University of Tokyo \& Statistical Mathematics Unit, RIKEN Center for Brain Science, e-mail: \texttt{matsuda@mist.i.u-tokyo.ac.jp}}}
	
	\date{}
	
	\maketitle
	
	\begin{abstract}
		The problem of predicting unobserved entries in a binary matrix, known as 1-bit matrix completion, has found diverse applications in fields such as recommendation systems.
		In this study, we develop an empirical Bayes method for 1-bit matrix completion motivated by the Efron--Morris estimator, a matrix generalization of the James--Stein estimator that shrinks singular values toward zero. 
		The proposed method exploits the underlying low-rank structure of binary matrices, drawing parallels with multidimensional item response theory. 		
		Simulation studies and real-data applications demonstrate that the proposed method achieves competitive predictive accuracy and favorable predictive calibration.
	\end{abstract}
	
	\section{Introduction}	
Matrix completion is a fundamental problem in machine learning, where the objective is to recover missing entries of a partially observed matrix. 
A prominent example is the Netflix Prize \citep{Bennett}, which involved predicting a matrix of movie ratings by users for recommendation purposes.
Beyond recommendation, matrix completion has recently found applications in causal inference for panel data \citep{Athey}.
A standard assumption in matrix completion is that the underlying matrix is approximately low-rank, reflecting a few latent factors that govern interactions between rows and columns. 
A substantial body of work has established theoretical guarantees and developed efficient algorithms for matrix completion \citep{Cai,Candes,Keshavan,Mazumder,Recht}, predominantly focusing on cases where the observed entries are continuous-valued.

In many applications, however, observations are not continuous-valued but binary. 
For instance, implicit feedback in recommender systems only records whether a user interacted with an item, while social networks encode the presence or absence of links. 
Similarly, preference data often consist of binary comparisons. 
Standard matrix completion algorithms designed for continuous-valued measurements are ill-suited to such data, as they fail to account for the quantized nature of the observations. 
This limitation has motivated the development of specialized techniques known as \emph{1-bit matrix completion} \citep{Cai13, Davenport, Liu}. 
Existing approaches in this domain primarily rely on restricted maximum likelihood estimation, where low-rankness is enforced through explicit constraints on the underlying matrix.
However, such methods often require tedious hyperparameter tuning and lack a natural framework for uncertainty quantification.

Empirical Bayes is a flexible statistical framework that performs Bayesian inference using a prior distribution estimated from the data \citep{Robbins}. 
This data-driven approach facilitates adaptive inference that effectively captures the latent structures inherent in the data.
The efficacy of this paradigm has been demonstrated across a broad range of statistical procedures \citep{Efron}. 
A prominent example is the James–Stein estimator \citep{Stein74}, an empirical Bayes estimate of a normal mean vector with a rigorously established decision-theoretic advantage, widely known as Stein's paradox.
This estimator was extended to a normal mean matrix by \cite{Efron72}, and its theoretical properties have been extensively investigated in recent years \citep{Matsuda15,Matsuda22,Matsuda24}.
These studies demonstrate that the Efron--Morris estimator works particularly well when the unknown matrix is approximately low-rank due to its singular value shrinkage property.

In this study, we develop an empirical Bayes algorithm for 1-bit matrix completion.
The proposed method is based on a hierarchical model motivated from the Efron--Morris estimator.
It estimates the hyperparameter by using the Monte Carlo EM algorithm \citep{Wei} and predicts unobserved entries via Bayesian predictive distributions.
This procedure does not require heuristic parameter tuning other than tolerance, and naturally accounts for the parameter uncertainty in its predictions.
Simulation results and real-data applications demonstrate that the proposed method performs well in terms of both prediction accuracy and uncertainty quantification.
The MATLAB code used in our experiments is available at \url{https://github.com/takeru-matsuda/EB_1bit_matrix_completion}.

This paper is organized as follows.
Section~\ref{sec_back} introduces the formulation of 1-bit matrix completion and briefly reviews previous studies.
Section~\ref{sec_prop} provides the details of the proposed EB algorithm.
Sections~\ref{sec_sim} and \ref{sec_real} present the results of the simulation and real-data applications, respectively.
Section~\ref{sec_concl} gives concluding remarks.

\section{1-bit matrix completion}\label{sec_back}
Suppose that we have a partially observed binary matrix $Y \in \{0,1\}^{p \times q}$.
We assume that each entry $Y_{ij}$ of $Y$ is an independent Bernoulli random variable with probability $f(M_{ij})$:
\begin{align}
	Y_{ij} \mid M \sim {\rm Bernoulli} (f(M_{ij})), \quad (i,j) \in \Omega, \label{obs_model}
\end{align}
where $M \in \mathbb{R}^{p \times q}$ is a latent matrix and $f$ is a link function.
In this study, we mainly consider the probit link function
\begin{align}
	f(m) = \int_{-\infty}^m \frac{1}{\sqrt{2 \pi}} \exp \left( -\frac{t^2}{2} \right) {\rm d} t. \label{probit}
\end{align}
Other options of $f$ include the logistic link function.

Let $\Omega \subset \{1, \cdots, p \} \times \{ 1, \cdots, q \}$ be the set of indices of the observed entries and $Y_\Omega=(Y_{ij} \mid (i,j) \in \Omega)$ be the resulting observations of $Y$. 
The goal of 1-bit matrix completion is to predict the unobserved entries of $Y$ based on $Y_\Omega$.
As in the case of continuous-valued observations, the key assumption is that $M$ is approximately low-rank.
Such a low-rank structure naturally arises in latent factor models such as multidimensional item response theory \citep{Reckase}.
For illustration, consider a model without item intercepts, in which
\[
M = A^{\top} B, \quad A = [a_1, \dots, a_p], \quad B = [b_1, \dots, b_q],
\]
where $a_i\in\mathbb{R}^r$ represents the latent abilities of person $i$
and $b_j\in\mathbb{R}^r$ is the loading vector for item $j$.
Here, $r$ denotes the latent dimensionality of abilities, and $\mathrm{rank}(M)\leq r$.
Similar structures also appear in logistic matrix factorization \citep{Johnson} and PCA for binary data \citep{de Leeuw}.

Thus, 1-bit matrix completion is formulated as a constrained maximum likelihood estimation problem:
\begin{align*}
	\underset{{M}}{{\rm maximize}} & \quad \sum_{(i,j) \in \Omega} ( Y_{ij} \log f(M_{ij}) + (1-Y_{ij}) \log (1-f(M_{ij})) ), \\
	{\rm subject\ to} & \quad M \in \mathcal{C}, 
\end{align*}
where the feasible set $\mathcal{C}$ encodes the approximate low-rank structure of $M$.
For example, \cite{Davenport} adopted the trace-norm constraint\footnote{Although they also used the sup-norm constraint, they state that trace-norm constraint alone already works well in practice.}:
\begin{align*}
	\mathcal{C} = \{ M \mid \| M \|_\ast \leq C \}, \quad \| M \|_\ast = \sum_{k=1}^q \sigma_k(M),
\end{align*}
where $\sigma_k(M)$ is the $k$-th singular value of $M$.
It can be viewed as a convex relaxation of the rank constraint.
Other options include explicit rank constraint \citep{Bhaskar,Ni,Liu} and the max-norm constraint \citep{Cai13}. 

Recently, \cite{Liu} developed an efficient Majorization--Minimization algorithm using the Gauss--Newton method, referred to as MMGN. 
Based on a comprehensive comparison of existing algorithms, they concluded that MMGN achieves comparable estimation accuracy to established methods while being computationally more efficient. 
From a theoretical perspective, \cite{Davenport} established the near-optimality of their method with respect to the Frobenius norm and the Hellinger distance, while \cite{Alquier} and \cite{Cottet} provided theoretical analyses of Bayesian approaches to 1-bit matrix completion.

\section{Proposed method}\label{sec_prop}
In this section, we develop an empirical Bayes (EB) method for 1-bit matrix completion.
As explained in the previous section, 1-bit matrix completion is formulated as estimation of $M$ from the observed entries $Y_{\Omega}$ of $Y$ under the model \eqref{obs_model}.
Instead of a constrained maximum likelihood, we adopt an empirical Bayes approach by introducing an independent Gaussian prior on each row $m_1,\dots,m_p$ of $M$:
\begin{align}
	m_i &\sim {\rm N}_q (0, \Sigma), \quad i=1,\dots,p. \label{prior}
\end{align}
This prior specification is motivated from the singular value shrinkage estimator by \cite{Efron72}, as detailed in Section~\ref{sec_EfronMorris}.

The proposed method proceeds in two steps.
First, the hyperparameter $\Sigma$ in the prior \eqref{prior} is estimated by maximizing the marginal likelihood:
\begin{align*}
	\hat{\Sigma} = \argmax_{\Sigma} \log p(Y_{\Omega} \mid \Sigma).
\end{align*}
This maximization is executed via the Monte Carlo EM algorithm \citep{Wei} described in Section~\ref{sec_MCEM}.
Next, for each unobserved entry $Y_{ij}$ with $(i,j) \not\in \Omega$, we derive the Bayesian predictive distribution:
\begin{align*}
	p(Y_{ij} \mid Y_{\Omega}) &= \int p(Y_{ij} \mid M) \pi(M \mid Y_{\Omega}; \Sigma) {\rm d} M,
\end{align*}
where $\pi(M \mid Y_{\Omega}; \Sigma)$ is the posterior of $M$.
By integrating over the posterior, it explicitly accounts for the estimation uncertainty in $M$, providing more reliable predictions than point estimation approaches.
The detail is given in Section~\ref{sec_BP}.
While the prior \eqref{prior} assumes a zero mean for each row, Section~\ref{sec_hetero} extends this to a general mean to incorporate column-wise heterogeneity.

\subsection{Monte Carlo EM}\label{sec_MCEM}
For estimation of the hyperparameter $\Sigma$ in the prior \eqref{prior}, we employ the Monte Carlo Expectation--Maximization (MCEM) algorithm \citep{Wei}, which is a stochastic extension of the classical EM algorithm \citep{EMalgo} that approximates the intractable expectations in the E-step using MCMC samples.

The algorithm iterates between the E-step and the M-step. 
In the E-step, we obtain samples from the posterior $\pi(M \mid Y_{\Omega}; \Sigma_t)$ using Markov chain Monte Carlo (MCMC). 
In the case of the probit link \eqref{probit}, we utilize the data augmentation technique of \cite{Chib} to derive a Gibbs sampler in closed form\footnote{In the case of the logit link function, the P\'olya--Gamma augmentation technique \citep{PSW} can be used to efficiently sample from the posterior.}.
Specifically, we introduce a latent Gaussian matrix $Z = (z_1,\dots,z_p)^{\top} \in \mathbb{R}^{p \times q}$ such that
\begin{align*}
	z_i \mid M \sim {\rm N}_q(m_i, I_q), \quad i = 1,\dots,p,
\end{align*}
where each $z_i$ is independent, and assume that the observed binary matrix $Y$ is determined by the sign of $Z$:
\begin{align*}
	Y_{ij} = 
	\begin{cases}
		1 & \mathrm{if}\ Z_{ij} \geq 0, \\
		0 & \mathrm{if}\ Z_{ij} < 0.
	\end{cases}
\end{align*}
In Gibbs sampling, we alternately update $M$ and $Z$ from their full conditional distributions $p(M \mid Z,Y_{\Omega}; \Sigma_t)$ and $p(Z \mid M, Y_{\Omega}; \Sigma_t)$, respectively.
From
\[
p(M \mid Z,Y_{\Omega}; \Sigma_t) \propto \pi(M; \Sigma_t) p(Z \mid M),
\]
the full conditional distribution of $M= (m_1,\dots,m_p)^{\top}$ is independent across rows and given by
\[
m_i \mid Z,Y_{\Omega}; \Sigma_t \sim {\rm N}_{q} ((I_q+\Sigma_t^{-1})^{-1}z_i, (I_q+\Sigma_t^{-1})^{-1}), \quad i = 1,\dots,p.
\]
Similarly, the full conditional distribution of $Z$ is independent across entries and given by
\begin{align}
	Z_{ij} \mid M, Y_{\Omega}; \Sigma_t \sim \begin{cases} {\rm N}(M_{ij}, 1) & \mathrm{for}\ (i,j) \not\in \Omega, \\ {\rm N}_+(M_{ij}, 1) & \mathrm{for}\ (i,j) \in \Omega, \ Y_{ij} = 1, \\ {\rm N}_-(M_{ij}, 1) & \mathrm{for}\ (i,j) \in \Omega, \ Y_{ij} = 0, \\ \end{cases} \label{z_gibbs}
\end{align}
where ${\rm N}_+(M_{ij},1)$ and ${\rm N}_-(M_{ij},1)$ denote the normal distribution ${\rm N}(M_{ij},1)$ truncated to the positive and negative intervals, respectively.
Sampling from these truncated normal distributions can be efficiently performed using the inverse transform method.
In this manner, we obtain $N=100$ posterior samples\footnote{In a sensitivity analysis with $p=1000$ and $q\in\{100,500,1000\}$, increasing the Monte Carlo sample size from $N=100$ to $N=500$ did not consistently improve predictive accuracy. Thus, we retained $N=100$ in our experiments.} of $M$ and $Z$.

In the M-step, we update the parameter estimate by
\begin{align}
	\Sigma_{t+1} =  {\rm E}_{M \mid Y_{\Omega},\Sigma_t} \left[ \frac{1}{p} \sum_{i=1}^p m_i m_i^{\top} \right]. \label{Mstep}
\end{align}
See the Appendix for its derivation.
We approximate this expectation by Monte Carlo using the posterior samples from the Gibbs sampler above.

We iterate the above process until convergence. 
Since the inherent Monte Carlo error in the E-step makes it difficult to apply a standard stopping rule \citep{Levine}, we used a fixed number of 20 iterations for our experiments in Sections~\ref{sec_sim} and \ref{sec_real}.

\subsection{Bayesian prediction of unobserved entries}\label{sec_BP}
After estimating the hyperparameter $\Sigma$ by Monte Carlo EM, we predict the unobserved entries of $Y$ via Bayesian predictive distributions.
For an unobserved entry $Y_{ij}$ of $Y$ with $(i,j) \not\in \Omega$, its predictive probability is given by
\begin{align*}
	\Pr [Y_{ij}=1 \mid Y_{\Omega}; \hat{\Sigma}] 	&= \int f(M_{ij}) \pi(M \mid Y_{\Omega}; \hat{\Sigma}) {\rm d} M.
\end{align*}
We approximate it using posterior samples  $M^{(1)},\dots,M^{(N)}$ from $\pi(M \mid Y_{\Omega}; \hat{\Sigma})$:
\begin{align*}
	\Pr [Y_{ij}=1 \mid Y_{\Omega}; \hat{\Sigma}]  \approx \frac{1}{N} \sum_{t=1}^N f(M_{ij}^{(t)}).
\end{align*}
Note that the posterior samples are generated by MCMC during the E-step of Monte Carlo EM, as explained in the previous subsection.
This process yields a predictive probability for each unobserved entry of $Y$.
In real-data applications in Section~\ref{sec_real}, we will verify that these predictive probabilities are well-calibrated, providing a reliable uncertainty quantification.

Beyond marginal predictions, the proposed method can also provide joint predictive distributions for multiple unobserved entries.
For example, for two unobserved entries $Y_{ij}$ and $Y_{kl}$, their Bayesian predictive distribution is given by
\begin{align*}
	\Pr [Y_{ij}=1,Y_{kl}=1 \mid Y_{\Omega}] 
	&= \int f(M_{ij}) f(M_{kl}) \pi(M \mid Y_{\Omega}) {\rm d} M,
\end{align*}
which can be approximated as
\begin{align*}
	\Pr [Y_{ij}=1,Y_{kl}=1 \mid Y_{\Omega}] \approx \frac{1}{N} \sum_{t=1}^N f(M_{ij}^{(t)}) f(M_{kl}^{(t)}).
\end{align*}
Such information regarding the dependence between entries can be useful in downstream tasks.

\subsection{Relation to the Efron--Morris estimator}\label{sec_EfronMorris}
The proposed method is motivated by an empirical Bayes estimator for a normal mean matrix by \cite{Efron72}.
Suppose that we have a matrix observation $Y \in \mathbb{R}^{p \times q}$, where each entry of $Y$ is independently distributed as $Y_{ij} \sim {\rm N} (M_{ij},1)$.
The Efron--Morris estimator of $M$ is given by
\begin{align}
	\hat{M}_{{\rm EM}} = Y ( I_q-(p-q-1) (Y^{\top} Y)^{-1} ), \label{EM_estimator}
\end{align}
which reduces to the James--Stein estimator \citep{Stein74} when $q=1$.
When $p-q-1>0$, the Efron--Morris estimator is minimax and dominates the maximum likelihood estimator $\hat{M}=Y$ under the Frobenius loss:
\begin{align*}
	{\rm E}_M [ \| \hat{M}_{{\rm EM}}-M \|_{\mathrm{F}}^2 ] \leq pq.
\end{align*}
Recently, \cite{Matsuda22} showed that the Efron--Morris estimator dominates the maximum likelihood estimator even under the matrix quadratic loss:
\begin{align*}
	{\rm E} [(\hat{M}_{{\rm EM}}-M)^{\top} (\hat{M}_{{\rm EM}}-M)] \preceq p I_q,
\end{align*}
and \cite{Matsuda24} applied it to derive an adaptive minimax estimator in nonparametric estimation.
These studies demonstrate that the Efron--Morris estimator works particularly well when the unknown matrix is approximately low-rank due to its singular value shrinkage property.

The Efron--Morris estimator \eqref{EM_estimator} is derived as an empirical Bayes estimator by considering the Gaussian prior on $M= (m_1,\dots,m_p)^{\top}$ given by
\begin{equation}
	m_i \sim {\rm N}_{q} (0, \Sigma), \quad i=1,\dots,p, \label{gauss_prior}
\end{equation}
where $m_1,\dots,m_p$ are independent.
For this prior, the Bayes estimator (posterior mean) of $M$ is given by
\begin{align}
	\hat{M}=Y (I_q-(I_q+\Sigma)^{-1}). \label{Best}
\end{align}
Now, the hyperparameter $\Sigma$ is estimated from $Y$ by moment matching.
Since the marginal distribution of $Y=(y_1,\dots,y_p)^{\top}$ is independent across rows and given by $y_i \sim {\rm N}_{q} (0, I_q+\Sigma)$ for $i=1,\dots,p$, 
the marginal distribution of $Y^{\top} Y$ is $Y^{\top} Y \sim W_q (p, I_q+\Sigma)$, the Wishart distribution with $p$ degrees of freedom.
Thus, ${\rm E} [(Y^{\top} Y)^{-1}] = (p-q-1)^{-1} (I_q+\Sigma)^{-1}$ and it gives the moment estimate of $\Sigma$ by $(I_q+\hat{\Sigma})^{-1}=(p-q-1) (Y^{\top} Y)^{-1}$.
By substituting this into \eqref{Best}, the Efron--Morris estimator $\hat{M}_{{\rm EM}}$ in \eqref{EM_estimator} is obtained.

As the Efron--Morris estimator works well for low-rank matrices, empirical Bayes methods with Gaussian priors of the form \eqref{gauss_prior} are expected to utilize (approximate) low-rankness of $M$ effectively.
Based on this idea, empirical Bayes matrix completion algorithms have been developed for continuous-valued \citep{Matsuda19} and count-valued \citep{LMK} matrix data.
Recently, \cite{Chakraborty} developed an empirical Bayes method for data integration in a similar spirit.

Independence across rows does not preclude a shared low-rank structure.
For example, if $m_i=Ba_i$, where $B\in\mathbb{R}^{q\times r}$ is fixed and
$a_i{\sim}\mathrm{N}_r(0,I_r)$ independently, then the rows are
independent with covariance $\Sigma=BB^\top$, while
$\operatorname{rank}(M)\le r$.
More generally, a covariance matrix whose variation is concentrated in
a few eigendirections allows the rows to concentrate near a common
low-dimensional subspace.

\subsection{Extension to column-wise heterogeneity}\label{sec_hetero}
In practice, observations can exhibit significant heterogeneity across columns; for instance, some columns may be dominated by zeros, while others consist primarily of ones. 
Here, we extend the proposed method to explicitly account for such column-wise inhomogeneity. 
In Section~\ref{sec_sim}, we demonstrate through numerical experiments that this extension achieves superior prediction accuracy in highly heterogeneous settings.

We extend the prior \eqref{prior} to
\begin{align}
	m_i &\sim {\rm N}_q (\mu, \Sigma), \quad i=1,\dots,p, \label{prior2}
\end{align}
where $\mu=(\mu_1,\dots,\mu_q)^\top$ is shared across rows.
Then, we estimate both $\mu$ and $\Sigma$ by maximizing the marginal likelihood:
\begin{align*}
	(\hat{\mu},\hat{\Sigma}) = \argmax_{\mu,\Sigma} \log p(Y_{\Omega} \mid \mu, \Sigma).
\end{align*}
We employ the Monte Carlo EM algorithm like Section~\ref{sec_MCEM}.
Under the prior \eqref{prior2}, the full conditional distribution of $M$ becomes
\[
m_i \mid Z,Y_{\Omega}; \mu_t,\Sigma_t \sim {\rm N}_{q} ((I_q+\Sigma_t^{-1})^{-1} (z_i+\Sigma_t^{-1} \mu_t), (I_q+\Sigma_t^{-1})^{-1}), \quad i = 1,\dots,p,
\]
while that of $Z$ remains the same with \eqref{z_gibbs}.
The parameter update formula is given by
\begin{align*}
	\mu_{t+1} =  {\rm E}_{M \mid Y_{\Omega};\mu_t,\Sigma_t} \left[ \frac{1}{p} \sum_{i=1}^p m_i \right], \quad \Sigma_{t+1} =  {\rm E}_{M \mid Y_{\Omega};\mu_t,\Sigma_t} \left[ \frac{1}{p} \sum_{i=1}^p (m_i-\mu_{t+1}) (m_i-\mu_{t+1})^{\top} \right].
\end{align*}
After hyperparameter estimation, the Bayesian predictive distribution is computed via MCMC samples in a similar way to Section~\ref{sec_BP}.

\subsection{Remark on identifiability}
Under the zero-mean prior \eqref{prior}, the marginal prior on  $M_{ij}$ is ${\rm N}(0, \Sigma_{jj} )$, which implies that $Y_{ij} \sim {\rm Bernoulli}(1/2)$ marginally. 
In other words, the marginal distribution of each $Y_{ij}$ does not contain any information about $\Sigma$. 
Nevertheless, the joint distribution
of $Y_{i1},\dots,Y_{iq}$ contains some information about $\Sigma$.
To see this, consider the latent Gaussian representation
\[
z_i=m_i+\varepsilon_i,\qquad
\varepsilon_i\sim\mathrm{N}_q(0,I_q),
\]
where $\varepsilon_i$ is independent of $m_i$ and
$Y_{ij}=\mathbf{1}(Z_{ij}\ge0)$.
After integrating out $m_i$, we obtain
$z_i\sim\mathrm{N}_q(0,I_q+\Sigma)$.
Since positive coordinate-wise rescaling preserves signs, the joint
distribution of the binary responses depends on $\Sigma$ only through
the correlation matrix $R=D^{-1}(I_q+\Sigma)D^{-1}$, where $D=\mathrm{diag}(\sqrt{1+\Sigma_{11}},\dots,\sqrt{1+\Sigma_{qq}})$.
For example,
\[
\Pr[Y_{ij}=1,Y_{ik}=1\mid\Sigma]=\frac{1}{4}+\frac{1}{2\pi}\arcsin(R_{jk}), \quad j\ne k.
\]

Thus, the prior covariance $\Sigma$ itself is not identifiable.
Specifically, for any $c>1$, the covariance matrix $\Sigma'=c(I_q+\Sigma)-I_q$ induces the same correlation matrix $R$ and hence the same joint distribution of binary responses.
The Bayesian predictive distribution of unobserved entries is invariant among $\Sigma$ inducing the same $R$.
We retain $\Sigma$ as a redundant working parameter because this
parameterization yields a closed-form M-step.

Under the nonzero-mean prior \eqref{prior2}, the marginal distribution of $Y$
depends on $(\mu,\Sigma)$ only through $\alpha=D^{-1}\mu$ and $R=D^{-1}(I_q+\Sigma)D^{-1}$.
Thus, introducing a nonzero mean does not remove scale
non-identifiability. 

A related non-identifiability arises in empirical Bayes matrix completion for Gaussian observations \citep{Matsuda19}. 
As shown in Appendix B.3 of \cite{Matsuda19}, non-identifiability does not invalidate the EM likelihood-ascent argument. 

\section{Numerical experiments}\label{sec_sim}
In this section, we evaluate the performance of the proposed EB algorithms through numerical experiments.
In the following, we refer to the algorithms based on the priors \eqref{prior} and \eqref{prior2} as EB1 and EB2, respectively.
The MATLAB code used in our experiments is available at \url{https://github.com/takeru-matsuda/EB_1bit_matrix_completion}.

\subsection{Comparison methods}
We compare EB1 and EB2 with several state-of-the-art methods: MMGN \citep{Liu}, TraceNorm \citep{Davenport}, and MaxNorm \citep{Cai13}.
For the existing methods, we employed the MATLAB implementation provided by \cite{Liu} at \url{https://github.com/Xiaoqian-Liu/MMGN}.

The existing methods determine the matrix rank from the data, but they require the user to pre-specify a maximum rank. 
In the following, we set this maximum rank to 10. 
Notably, EB1 and EB2 do not require such manual constraints on the rank, which constitutes a practical advantage over the existing methods.

\subsection{Data generation}
Following previous studies \citep{Davenport,Cai13,Liu}, we generate synthetic data as follows using the probit link function \eqref{probit}:

\begin{enumerate}
	\item Sample $U_{ij} \sim {\rm Unif}[-1,1]$ independently for $i=1,\dots,p$ and $j=1,\dots,r$.
	
	\item Sample $V_{jk} \sim {\rm Unif}[-1,1]$ independently for $j=1,\dots,r$ and $k=1,\dots,q$.
	
	\item Compute $M = M_0+sUV$, where $M_0$ is a constant matrix (we set $M_0=O$ except for Table~\ref{tab_demo2}) and $s>0$.
	
	\item Randomly sample a subset $\Omega \subset \{ 1,\dots,p \} \times \{ 1,\dots,q \}$ of fixed size $|\Omega|$.
	
	\item Sample $Y_{ij} \sim {\rm Bernoulli} (f(M_{ij}))$ independently for $(i,j) \in \Omega$, where $f$ is the probit link function \eqref{probit}.
\end{enumerate}

The parameter $s>0$ controls the scale of the entries of $M$.
\cite{Liu} considered the probit link function with scale parameter $\sigma$ defined by
\[
f(m \mid \sigma^2) = \int_{-\infty}^m \frac{1}{\sqrt{2 \pi \sigma^2}} \exp \left( -\frac{t^2}{2 \sigma^2} \right) {\rm d} t,
\]
and demonstrated that the performance of 1-bit matrix completion methods depends on $\sigma^2$.
From $f(m \mid \sigma^2) = f (m/\sigma \mid 1)$, varying $s$ with $\sigma^2=1$ fixed is equivalent to varying $\sigma$.
Namely, they are in one-to-one correspondence via $s = 1/\sigma$.

\subsection{Performance metrics}
Each method outputs the predictive probability $p_{ij} = {\rm Pr}[Y_{ij}=1]$ for each unobserved entry $(i,j) \not \in \Omega$.
To evaluate predictive accuracy, we consider the average of the Kullback--Leibler divergence from the true probability $f(M_{ij})$ to the predictive probability $p_{ij}$ over all unobserved entries:
\begin{align*}
	\frac{1}{pq-|\Omega|} \sum_{(i,j) \not\in \Omega} \left( f(M_{ij}) \log \frac{f(M_{ij})}{p_{ij}} + (1-f(M_{ij})) \log \frac{1-f(M_{ij})}{1-p_{ij}} \right).
\end{align*}
This is a standard metric for evaluating predictive distributions \citep{Aitchison}.

In addition, following \cite{Liu}, we also consider the average of the Hellinger distance between the true probability $f(M_{ij})$ and the predictive probability $p_{ij}$ over all unobserved entries:
\begin{align*}
	\frac{1}{pq-|\Omega|} \sum_{(i,j) \not\in \Omega} \left( (\sqrt{p_{ij}}-\sqrt{f(M_{ij})})^2 + (\sqrt{1-p_{ij}}-\sqrt{1-f(M_{ij})})^2 \right).
\end{align*}

We found that MMGN sometimes outputs $p_{ij}=0$ or $p_{ij}=1$ (up to machine precision).
In such cases, the Kullback--Leibler divergence may become infinite, whereas the Hellinger distance remains finite.

\subsection{Results}
Table~\ref{tab_demo1} summarizes the performance metrics averaged over 100 independent replications when $p=1000$, $q=100$, $r=5$, $s=1$, $|\Omega|/pq=0.5$, and $M_0=O$.
Both EB1 and EB2 achieve the smallest errors in terms of both Kullback--Leibler divergence and Hellinger distance. 
While MMGN exhibits a smaller computation time than EB1 and EB2, it occasionally produces predictive probabilities of 0 or 1 (up to machine precision), causing the KL divergence to diverge.

Table~\ref{tab_demo2} presents the results under the same setting with Table~\ref{tab_demo1} except for $(M_0)_{ij} = a_j$, where $a_1,\dots,a_q$ are independent samples from the uniform distribution on $[-3,3]$.
Here, EB2 attains superior prediction accuracy compared to EB1.
This result demonstrates the effectiveness of the prior \eqref{prior2} in adaptively capturing heterogeneous column effects.

\begin{table}
	\renewcommand{\arraystretch}{1.3}
	\caption{Performance of 1-bit matrix completion algorithms for $p=1000$, $q=100$, $r=5$, $s=1$, and $|\Omega|/pq=0.5$. Computation time is given in seconds.}
	\label{tab_demo1}
	\centering
	\begin{tabular}{|c|c|c|c|}
		\hline
		& KL & Hellinger & time \\ \hline
		MMGN & $\infty$ & 0.054 & 14.39 \\ \hline
		TraceNorm & 0.067 & 0.032 & 61.54 \\ \hline
		MaxNorm & 0.075 & 0.042  & 50.77 \\ \hline
		EB1 & 0.061 & 0.031 & 44.70 \\ \hline
		EB2 & 0.062 & 0.032 & 44.83 \\ \hline
	\end{tabular}
\end{table}

\begin{table}
	\renewcommand{\arraystretch}{1.3}
	\caption{Performance of 1-bit matrix completion algorithms for $p=1000$, $q=100$, $r=5$, and $|\Omega|/pq=0.5$. Computation time is given in seconds.}
	\label{tab_demo2}
	\centering
	\begin{tabular}{|c|c|c|c|}
		\hline
		& KL & Hellinger & time \\ \hline
		MMGN & 0.087 & 0.045 & 24.74 \\ \hline
		TraceNorm & 0.079 & 0.041 & 240.60 \\ \hline
		MaxNorm & 0.107 & 0.073  & 1353.98 \\ \hline
		EB1 & 0.061 & 0.036 & 53.40 \\ \hline
		EB2 & 0.057 & 0.033 & 52.36 \\ \hline
	\end{tabular}
\end{table}

Now, we investigate how the performance of each algorithm depends on the matrix size $(p,q)$, the matrix rank $r$, the matrix scale $s$, and the proportion of observed entries $|\Omega|/(pq)$.
For these experiments, we set $M_0=O$ and report the results averaged over 100 independent replications.
We observed that the Kullback--Leibler divergence and the Hellinger distance exhibit similar trends, although the Kullback--Leibler divergence may diverge for MMGN in some cases.
Therefore, we focus on the Hellinger distance in the following.
Furthermore, since EB1 and EB2 yielded nearly identical performance in this homogeneous setting, we present only the results for EB1 for the sake of brevity.

Figure~\ref{fig_varying_p} plots the Hellinger distance and computation time  as functions of $p$, with $q=100$, $r=5$, $s=1$, and $|\Omega|/(pq) = 0.5$.
EB1 and TraceNorm achieve the best predictive accuracy across the tested values of $p$, while EB1 requires less computation time.
The Hellinger distance of EB1 decreases as $p$ increases.
This behavior is consistent with the favorable performance of the Efron--Morris estimator in large-$p$, low-rank settings \citep{Matsuda22,Matsuda24}.
Although EB1 is slower than MMGN, its runtime relative to MMGN decreases from approximately 3.0 at $p=500$ to 1.3 at $p=5000$.
These results highlight the effectiveness of EB1 for matrices with many rows and relatively few columns.

Figure~\ref{fig_varying_q} plots the Hellinger distance and
computation time as functions of $q$, with $p=1000$, $r=5$,
$s=1$, and $|\Omega|/(pq)=0.5$.
For larger $q$, MMGN and TraceNorm achieve the best predictive
accuracy, while MMGN requires substantially less computation time.
In contrast, the Hellinger distance of EB1 increases as $q$
increases.
Although EB1 remains faster than TraceNorm, it is slower than MMGN for large $q$.
These results indicate that EB1 is less effective when the number
of columns is comparable to the number of rows, both in predictive
accuracy and computational efficiency.

Figure~\ref{fig_varying_rho} plots the Hellinger distance and computation time as functions of $|\Omega|/(pq)$, with $p=1000$, $q=100$, $r=5$, and $s=1$.
Similarly to Figure~\ref{fig_varying_p}, EB1 achieves better accuracy than the other methods with computational cost comparable with MMGN.
The prediction accuracy improves as $| \Omega |/(pq)$ increases due to more information from observations.
Note that $| \Omega |/(pq)$ can be as small as 0.1 in applications of 1-bit matrix completion such as MovieLens dataset (Section~\ref{sec_real}).

Figure~\ref{fig_varying_rank} plots the Hellinger distance and computation time as functions of $r$, with $p=1000$, $q=100$, and $|\Omega|/pq = 0.5$.
While the prediction accuracy degrades with increasing rank for all methods, EB1 is less affected and outperforms the other methods for larger ranks.
The computation time of EB1 is nearly constant with respect to the rank.

Figure~\ref{fig_varying_s} plots the Hellinger distance and computation time as functions of $s$, with $p=1000$, $q=100$, $r=5$, and $|\Omega|/pq = 0.5$.
Note that $s$ corresponds to signal-to-noise ratio in 1-bit matrix completion \citep{Davenport}.
EB1 exhibits relatively stable predictive performance across a wide range of scales. 

Overall, EB1 combines high predictive accuracy with moderate computational cost when the number of rows is large relative to the number of columns.
Its predictive accuracy deteriorates less rapidly with increasing rank than that of the competing methods and remains relatively stable across a wide range of signal scales.
These results demonstrate the effectiveness of empirical Bayes singular value shrinkage in adapting to low-rank structure without manual rank selection, consistent with findings for Gaussian matrix completion \citep{Matsuda19}.

\begin{figure}
	\centering
	\begin{tikzpicture}
		\begin{axis}[
			xlabel={$p$},
			ylabel={Hellinger},
			width=6cm,
			height=6cm,
			grid=major,
			thick,
			legend to name=scalingplegend,
			legend columns=4,
			]
			\addplot[blue,mark=o,thick] coordinates {
				(500,0.06064974787)
				(1000,0.05484384104)
				(2000,0.04940221773)
				(5000,0.04451678508)
			};
			\addlegendentry{MMGN}
			\addplot[black,mark=square,thick] coordinates {
				(500,0.04328204470)
				(1000,0.03288778967)
				(2000,0.02757998724)
				(5000,0.02524775106)
			};
			\addlegendentry{TraceNorm}
			\addplot[green!60!black,mark=triangle,thick] coordinates {
				(500,0.04150429827)
				(1000,0.04223984403)
				(2000,0.04188216451)
				(5000,0.04206657779)
			};
			\addlegendentry{MaxNorm}
			\addplot[red,mark=*,very thick] coordinates {
				(500,0.04145929125)
				(1000,0.03194611931)
				(2000,0.02747863180)
				(5000,0.02531404535)
			};
			\addlegendentry{EB1}
		\end{axis}
	\end{tikzpicture}
	\begin{tikzpicture}
		\begin{axis}[
			xlabel={$p$},
			ylabel={time},
			width=6cm,
			height=6cm,
			grid=major,
			thick,
			]
			\addplot[blue,mark=o,thick] coordinates {
				(500,7.251506899)
				(1000,14.38899514)
				(2000,43.40521552)
				(5000,169.3534586)
			};
			\addplot[black,mark=square,thick] coordinates {
				(500,37.94721382)
				(1000,62.19716729)
				(2000,133.0740797)
				(5000,539.3217577)
			};
			\addplot[green!60!black,mark=triangle,thick] coordinates {
				(500,46.66077930)
				(1000,51.49330981)
				(2000,78.10168495)
				(5000,227.8336218)
			};
			\addplot[red,mark=*,very thick] coordinates {
				(500,21.72332845)
				(1000,41.19926719)
				(2000,81.07275650)
				(5000,214.6398649)
			};
		\end{axis}
	\end{tikzpicture}
	\par\vspace{2mm}
	\pgfplotslegendfromname{scalingplegend}
	\caption{Comparison of the Hellinger distance (left) and computation time (right) as functions of $p$ when $q=100$, $r=5$, $s=1$, and $|\Omega|/pq=0.5$. Results are averaged over 100 simulation replicates. Computation time is in seconds.}
	\label{fig_varying_p}
\end{figure}

\begin{figure}
	\centering
	\begin{tikzpicture}
		\begin{axis}[
			xlabel={$q$},
			ylabel={Hellinger},
			width=6cm,
			height=6cm,
			grid=major,
			thick,
			legend to name=scalingqlegend,
			legend columns=4,
			]
			\addplot[blue,mark=o,thick] coordinates {
				(100,0.05484384104)
				(200,0.01981282068)
				(500,0.008752237135)
				(1000,0.005533002542)
			};
			\addlegendentry{MMGN}
			\addplot[black,mark=square,thick] coordinates {
				(100,0.03288778967)
				(200,0.01653324213)
				(500,0.008209814934)
				(1000,0.005360355104)
			};
			\addlegendentry{TraceNorm}
			\addplot[green!60!black,mark=triangle,thick] coordinates {
				(100,0.04223984403)
				(200,0.04050385519)
				(500,0.03905338671)
				(1000,0.03860091437)
			};
			\addlegendentry{MaxNorm}
			\addplot[red,mark=*,very thick] coordinates {
				(100,0.03194611931)
				(200,0.03012374525)
				(500,0.04431924898)
				(1000,0.06226726184)
			};
			\addlegendentry{EB1}
		\end{axis}
	\end{tikzpicture}
	\begin{tikzpicture}
		\begin{axis}[
			xlabel={$q$},
			ylabel={time},
			width=6cm,
			height=6cm,
			grid=major,
			thick,
			]
			\addplot[blue,mark=o,thick] coordinates {
				(100,55.49169988)
				(200,26.89483744)
				(500,21.98764153)
				(1000,48.59532324)
			};
			\addplot[black,mark=square,thick] coordinates {
				(100,67.05975861)
				(200,122.0013509)
				(500,447.1422593)
				(1000,2656.154562)
			};
			\addplot[green!60!black,mark=triangle,thick] coordinates {
				(100,77.42360677)
				(200,77.80739009)
				(500,215.3020776)
				(1000,616.6675472)
			};
			\addplot[red,mark=*,very thick] coordinates {
				(100,44.56545413)
				(200,91.11381581)
				(500,264.5997934)
				(1000,807.9206110)
			};
		\end{axis}
	\end{tikzpicture}
	\par\vspace{2mm}
	\pgfplotslegendfromname{scalingqlegend}
	\caption{Comparison of the Hellinger distance (left) and computation time (right) as functions of $q$ when $p=1000$, $r=5$, $s=1$, and $|\Omega|/pq=0.5$. Results are averaged over 100 simulation replicates. Computation time is in seconds.}
	\label{fig_varying_q}
\end{figure}

\begin{figure}
	\centering
	\begin{tikzpicture}
		\begin{axis}[
			xlabel={$|\Omega|/(pq)$},
			ylabel={Hellinger},
			width=6cm,
			height=6cm,
			grid=major,
			thick,
			legend to name=commonlegend,
			legend columns=4,
			]
			
			\addplot[
			color=blue,
			mark=o,
			thick
			]
			table [col sep=comma, x=rho, y=MMGN] {hellinger_unobs_varying_rho.csv};
			\addlegendentry{MMGN}
			
			\addplot[
			color=black,
			mark=square,
			thick
			]
			table [col sep=comma, x=rho, y=Trace] {hellinger_unobs_varying_rho.csv};
			\addlegendentry{TraceNorm}
			
			\addplot[
			color=green!60!black,
			mark=triangle,
			thick
			]
			table [col sep=comma, x=rho, y=Max] {hellinger_unobs_varying_rho.csv};
			\addlegendentry{MaxNorm}
			
			\addplot[
			color=red,
			mark=*,
			very thick
			]
			table [col sep=comma, x=rho, y=EB1] {hellinger_unobs_varying_rho.csv};
			\addlegendentry{EB1}
			
		\end{axis}
	\end{tikzpicture}
	\begin{tikzpicture}
		\begin{axis}[
			xlabel={$|\Omega|/(pq)$},
			ylabel={time},
			width=6cm,
			height=6cm,
			grid=major,
			thick,
			]
			
			\addplot[
			color=blue,
			mark=o,
			thick
			]
			table [col sep=comma, x=rho, y=MMGN] {time_varying_rho.csv};
			
			\addplot[
			color=black,
			mark=square,
			thick
			]
			table [col sep=comma, x=rho, y=Trace] {time_varying_rho.csv};
			
			\addplot[
			color=green!60!black,
			mark=triangle,
			thick
			]
			table [col sep=comma, x=rho, y=Max] {time_varying_rho.csv};
			
			\addplot[
			color=red,
			mark=*,
			very thick
			]
			table [col sep=comma, x=rho, y=EB1] {time_varying_rho.csv};
			
		\end{axis}
	\end{tikzpicture}
	\vspace{2mm}
	\pgfplotslegendfromname{commonlegend}
	\caption{Comparison of the Hellinger distance (left) and computation time (right) as functions of $|\Omega|/pq$ when $p=1000$, $q=100$, $r=5$, and $s=1$.}
	\label{fig_varying_rho}
\end{figure}

\begin{figure}
	\centering
	\begin{tikzpicture}
		\begin{axis}[
			xlabel={$r$},
			ylabel={Hellinger},
			width=6cm,
			height=6cm,
			grid=major,
			thick,
			legend to name=commonlegend,
			legend columns=4,
			]
			
			\addplot[
			color=blue,
			mark=o,
			thick
			]
			table [col sep=comma, x=rank, y=MMGN] {hellinger_unobs_varying_rank.csv};
			\addlegendentry{MMGN}
			
			\addplot[
			color=black,
			mark=square,
			thick
			]
			table [col sep=comma, x=rank, y=Trace] {hellinger_unobs_varying_rank.csv};
			\addlegendentry{TraceNorm}
			
			\addplot[
			color=green!60!black,
			mark=triangle,
			thick
			]
			table [col sep=comma, x=rank, y=Max] {hellinger_unobs_varying_rank.csv};
			\addlegendentry{MaxNorm}
			
			\addplot[
			color=red,
			mark=*,
			very thick
			]
			table [col sep=comma, x=rank, y=EB1] {hellinger_unobs_varying_rank.csv};
			\addlegendentry{EB1}
			
		\end{axis}
	\end{tikzpicture}
	\begin{tikzpicture}
		\begin{axis}[
			xlabel={$r$},
			ylabel={time},
			width=6cm,
			height=6cm,
			grid=major,
			thick,
			]
			
			\addplot[
			color=blue,
			mark=o,
			thick
			]
			table [col sep=comma, x=rank, y=MMGN] {time_varying_rank.csv};
			
			\addplot[
			color=black,
			mark=square,
			thick
			]
			table [col sep=comma, x=rank, y=Trace] {time_varying_rank.csv};
			
			\addplot[
			color=green!60!black,
			mark=triangle,
			thick
			]
			table [col sep=comma, x=rank, y=Max] {time_varying_rank.csv};
			
			\addplot[
			color=red,
			mark=*,
			very thick
			]
			table [col sep=comma, x=rank, y=EB1] {time_varying_rank.csv};
			
		\end{axis}
	\end{tikzpicture}
	\vspace{2mm}
	\pgfplotslegendfromname{commonlegend}
	\caption{Comparison of the Hellinger distance (left) and computation time (right) as functions of $r$ when $p=1000$, $q=100$, $s=1$, and $|\Omega|/pq=0.5$.}
	\label{fig_varying_rank}
\end{figure}

\begin{figure}
	\centering
	\begin{tikzpicture}
		\begin{axis}[
			xlabel={$s$},
			ylabel={Hellinger},
			width=6cm,
			height=6cm,
			grid=major,
			thick,
			legend to name=commonlegend,
			legend columns=4,
			]
			
			\addplot[
			color=blue,
			mark=o,
			thick
			]
			table [col sep=comma, x=s, y=MMGN] {hellinger_unobs_varying_s.csv};
			\addlegendentry{MMGN}
			
			\addplot[
			color=black,
			mark=square,
			thick
			]
			table [col sep=comma, x=s, y=Trace] {hellinger_unobs_varying_s.csv};
			\addlegendentry{TraceNorm}
			
			\addplot[
			color=green!60!black,
			mark=triangle,
			thick
			]
			table [col sep=comma, x=s, y=Max] {hellinger_unobs_varying_s.csv};
			\addlegendentry{MaxNorm}
			
			\addplot[
			color=red,
			mark=*,
			very thick
			]
			table [col sep=comma, x=s, y=EB1] {hellinger_unobs_varying_s.csv};
			\addlegendentry{EB1}
			
		\end{axis}
	\end{tikzpicture}
	\begin{tikzpicture}
		\begin{axis}[
			xlabel={$s$},
			ylabel={time},
			width=6cm,
			height=6cm,
			grid=major,
			thick,
			legend style={
				at={(1.02,1)},
				anchor=north west,
				draw=none
			},
			]
			
			\addplot[
			color=blue,
			mark=o,
			thick
			]
			table [col sep=comma, x=s, y=MMGN] {time_varying_s.csv};
			
			\addplot[
			color=black,
			mark=square,
			thick
			]
			table [col sep=comma, x=s, y=Trace] {time_varying_s.csv};
			
			\addplot[
			color=green!60!black,
			mark=triangle,
			thick
			]
			table [col sep=comma, x=s, y=Max] {time_varying_s.csv};
			
			\addplot[
			color=red,
			mark=*,
			very thick
			]
			table [col sep=comma, x=s, y=EB1] {time_varying_s.csv};
			
		\end{axis}
	\end{tikzpicture}
	\vspace{2mm}
	\pgfplotslegendfromname{commonlegend}
	\caption{Comparison of the Hellinger distance (left) and computation time (right) as functions of $s$ when $p=1000$, $q=100$, $r=5$, and $|\Omega|/pq=0.5$.}
	\label{fig_varying_s}
\end{figure}

\section{Applications to real data}\label{sec_real}
In this section, we examine the performance of the proposed EB algorithms on real data.
The MATLAB code used in our experiments is available at \url{https://github.com/takeru-matsuda/EB_1bit_matrix_completion}.

\subsection{Performance metrics}
To evaluate the performance of the 1-bit matrix completion algorithms, we employ a masking procedure where a subset of the observed entries is hidden and then predicted.
Let $\Omega_0 \subset \{ 1,\dots,p \} \times \{ 1,\dots,q \}$ denote the set of observed entries in a given data.
We randomly partition $\Omega_0$ into two disjoint subsets, $\Omega$ and $\Omega_0 \setminus \Omega$, and mask the entries $Y_{ij}$ for $(i,j) \in \Omega_0 \setminus \Omega$.
We then apply the 1-bit matrix completion algorithms to predict the masked entries based on the observed entries $Y_{\Omega}$. 

Let $p_{ij} = {\rm Pr}[Y_{ij}=1 \mid Y_\Omega]$ denote the predictive probability for $(i,j) \in \Omega_0 \setminus \Omega$.
We evaluate predictive accuracy in two ways.
First, we convert $p_{ij}$ to point predictions using a threshold of 0.5:
\begin{align*}
	\hat{Y}_{ij} = \begin{cases} 1 & (p_{ij} \geq 0.5) \\ 0 & (p_{ij} < 0.5) \end{cases},
\end{align*}
and compute the classification accuracy:
\begin{align*}
	{\rm Accuracy} = \frac{1}{|\Omega_0 \setminus \Omega|} \sum_{(i,j) \in \Omega_0 \setminus \Omega} \mathbf{1}(\hat{Y}_{ij}=Y_{ij}).
\end{align*}
The second metric is the average cross-entropy loss over $(i,j) \in \Omega_0 \setminus \Omega$:
\begin{align*}
	\text{Cross-Entropy} = - \frac{1}{|\Omega_0 \setminus \Omega|} \sum_{(i,j) \in \Omega_0 \setminus \Omega} ( Y_{ij} \log p_{ij} + (1-Y_{ij}) \log (1-p_{ij}) ).
\end{align*}
Its expectation equals the average Kullback--Leibler divergence plus the average entropy of the true Bernoulli distributions, which does not depend on the prediction method.

We also assess the calibration performance of each method, where calibration refers to the agreement between predicted probabilities and observed empirical frequencies.
To the best of our knowledge, calibration performance of 1-bit matrix completion algorithms on real data has not been studied well.
We partition the predicted probabilities into $K$ bins, $I_k = [(k-1)/K, k/K)$ for $k=1,\dots,K-1$ and $I_K = [(K-1)/K, 1]$. 
Let 
\begin{align*}
	B_k = \{(i,j) \in \Omega_0 \setminus \Omega \mid p_{ij} \in I_k\}
\end{align*}
be the set of entries whose predictive probabilities fall into bin $I_k$.
We then evaluate uncertainty quantification via the expected calibration error (ECE) \citep{Dawid,Guo}:
\begin{align*}
	{\rm ECE} = \sum_{k=1}^K \frac{|B_k|}{|\Omega_0 \setminus \Omega|} \left|{\rm pred}(B_k)-{\rm real}(B_k)\right|,
\end{align*}
where 
\begin{align*}
	{\rm pred}(B_k) = \frac{1}{|B_k|} \sum_{(i,j) \in B_k} p_{ij}, 
	\quad 
	{\rm real}(B_k) = \frac{1}{|B_k|} \sum_{(i,j) \in B_k} Y_{ij}.
\end{align*}
The ECE quantifies the discrepancy between predicted probabilities and observed frequencies; a smaller ECE value indicates superior calibration. 
We set $K=10$ in the following experiments.

\subsection{Jester dataset}
We first evaluate the algorithms using the Jester dataset \citep{Goldberg} from \url{https://eigentaste.berkeley.edu/dataset/}.
This dataset contains $1{,}810{,}455$ ratings from $p=24{,}983$ users on $q=100$ jokes. 
The original ratings are continuous-valued, ranging from $-10.00$ to $10.00$. 
We binarize these ratings by mapping non-negative values to 1 and negative values to 0. 
For our experiments, we randomly mask $50\%$ of the observed entries as a test set and apply 1-bit matrix completion algorithms to predict these masked values.

Table~\ref{tab_jester} presents the classification accuracy, cross-entropy loss, ECE, and computation time for each algorithm. 
In this dataset, the EB methods outperform all existing methods in terms of both classification accuracy and cross-entropy loss. 
Specifically, EB2 achieves the highest accuracy and the smallest cross-entropy loss among the tested algorithms.
Regarding uncertainty quantification, the ECE values of EB1 and EB2 are remarkably small (both $<0.02$), indicating superior calibration compared to the other methods, including TraceNorm and MMGN. 
This suggests that EB methods provide well-calibrated confidence estimates, which is crucial for reliable decision-making under uncertainty \citep{Guo}.
Figure~\ref{fig_jester} shows the reliability diagrams, where the empirical frequency closely follows the prediction probability for the EB methods, further confirming their excellent calibration properties. 
While TraceNorm shows competitive accuracy, its computational cost is significantly larger than that of the EB methods. 

\begin{table}[tbp]
	\centering
	\renewcommand{\arraystretch}{1.3}
	\caption{Performance of 1-bit matrix completion algorithms on the
		Jester dataset. Computation time is given in seconds.}
	\label{tab_jester}
	\begin{tabular}{|c|c|c|c|c|}
		\hline
		& Accuracy & Cross-entropy & ECE & Time \\
		\hline
		MMGN      & 0.7148 & 0.5676 & 0.0319 & 222.38 \\
		\hline
		TraceNorm & 0.7159 & 0.5499 & 0.0262 & 1472.37 \\
		\hline
		MaxNorm   & 0.7065 & 0.6203 & 0.1313 & 155.02 \\
		\hline
		EB1       & 0.7253 & 0.5393 & 0.0197 & 300.90 \\
		\hline
		EB2       & 0.7301 & 0.5329 & 0.0193 & 311.82 \\
		\hline
	\end{tabular}
\end{table}

\begin{figure}[tbp]
	\centering
	\begin{tikzpicture}
		\begin{axis}[
			width=0.65\linewidth,height=0.65\linewidth,
			xmin=0,xmax=1,ymin=0,ymax=1,
			xtick={0,0.2,0.4,0.6,0.8,1},ytick={0,0.2,0.4,0.6,0.8,1},
			tick label style={font=\scriptsize},
			label style={font=\small},title style={font=\small},
			xlabel={Mean predicted probability},
			ylabel={Observed proportion of ones},
			grid=major,grid style={gray!20},
			legend columns=5,
			legend to name=jesterupdatedlegend,
			]
			\addplot[black,dashed,forget plot] coordinates {(0,0) (1,1)};
			\addplot[color=blue,mark=o,thick,mark size=2pt] coordinates {
				(0.03840964,0.12732657)
				(0.15211039,0.22989062)
				(0.25292824,0.30653580)
				(0.35265300,0.38456775)
				(0.45122759,0.46596319)
				(0.55024330,0.55127167)
				(0.64985457,0.64190167)
				(0.74995993,0.72881181)
				(0.85022216,0.81452346)
				(0.96255667,0.91470340)
			};
			\addlegendentry{MMGN}
			\addplot[color=black,mark=square,thick,mark size=2pt] coordinates {
				(0.06390036,0.09239194)
				(0.15254139,0.19741969)
				(0.25214705,0.29499557)
				(0.35137767,0.37853322)
				(0.45060350,0.45799226)
				(0.55051014,0.54017473)
				(0.65061254,0.62320613)
				(0.75085597,0.71672930)
				(0.85161907,0.81867812)
				(0.94320503,0.92283709)
			};
			\addlegendentry{TraceNorm}
			\addplot[color=green!60!black,mark=triangle,thick,mark size=2pt] coordinates {
				(0.18894435,0.00000000)
				(0.27848660,0.08515882)
				(0.36628949,0.18667405)
				(0.45774601,0.40320362)
				(0.54804467,0.67209272)
				(0.63853366,0.86864824)
				(0.72222888,0.93201238)
				(0.80929970,0.95312500)
			};
			\addlegendentry{MaxNorm}
			\addplot[color=red,mark=*,very thick,mark size=2pt] coordinates {
				(0.07311643,0.05633803)
				(0.15520395,0.13415935)
				(0.25349542,0.23215450)
				(0.35224072,0.34071920)
				(0.45115210,0.44843600)
				(0.55046598,0.56143762)
				(0.65022163,0.67348472)
				(0.74993055,0.78281601)
				(0.84970665,0.88299290)
				(0.93150367,0.94727773)
			};
			\addlegendentry{EB1}
			\addplot[color=magenta,mark=diamond*,very thick,dashed,mark size=2pt] coordinates {
				(0.07151426,0.05007880)
				(0.15584311,0.12229365)
				(0.25316392,0.22563286)
				(0.35189031,0.33937422)
				(0.45101383,0.44648698)
				(0.55058213,0.55677696)
				(0.65043250,0.66901968)
				(0.75042539,0.77824421)
				(0.84973028,0.88004044)
				(0.93072856,0.95144593)
			};
			\addlegendentry{EB2}
		\end{axis}
	\end{tikzpicture}
	\par\vspace{2mm}
	\pgfplotslegendfromname{jesterupdatedlegend}
	\caption{Reliability diagrams on the Jester dataset.
		Each point shows the mean predicted probability and the observed proportion of ones
		within a nonempty probability bin, evaluated on held-out observed entries.
		The dashed line indicates perfect calibration.}
	\label{fig_jester}
\end{figure}

\subsection{MovieLens 100K dataset}
To further investigate the robustness of the EB methods, we next consider the MovieLens 100K dataset from \url{https://grouplens.org/datasets/movielens/100k/}, a popular benchmark for matrix completion.
The dataset contains $10^5$ ratings from $q=943$ users across 1,682 movies, with each user providing at least 20 ratings on a scale from 1 to 5.
Following the preprocessing in \cite{Davenport}, we binarize the ratings by mapping 1--3 to 0 and 4--5 to 1. 
We select $p=1{,}349$ movies that received at least five ratings. 
The resulting data matrix $Y \in \{ 0,1 \}^{p \times q}$ contains $|\Omega_0|=99{,}287$ observed entries out of a total of $pq=1{,}272{,}107$. 
For our experiments, we randomly mask $50\%$ of the observed entries as a test set and apply 1-bit matrix completion algorithms to predict these masked values.

Table~\ref{tab_movie} summarizes the classification accuracy, cross-entropy loss, Expected Calibration Error (ECE), and computation time for each algorithm. 
EB1 and EB2 achieve classification accuracy and cross-entropy loss comparable to or better than existing methods. 
While MMGN yields a divergent cross-entropy loss ($\infty$) due to over-confident predictions near the boundaries, the EB methods maintain stable and small cross-entropy loss.
Regarding uncertainty quantification, the ECE of the EB methods is notably small ($<0.05$), which is comparable to MMGN and significantly better than MaxNorm. 
This calibration performance is further visualized in the reliability diagrams in Figure~\ref{fig_movie}, confirming that all algorithms except for MaxNorm exhibit favorable calibration properties.
In terms of computational efficiency, the EB methods provide a practical balance; they are significantly faster than TraceNorm while maintaining competitive predictive performance.

\begin{table}[tbp]
	\centering
	\renewcommand{\arraystretch}{1.3}
	\caption{Performance of 1-bit matrix completion algorithms on the MovieLens 100K dataset. Computation time is given in seconds.}
	\label{tab_movie}
	\begin{tabular}{|c|c|c|c|c|}
		\hline
		& Accuracy & Cross-entropy & ECE & Time \\
		\hline
		MMGN & 0.6598 & $\infty$ & 0.0299 & 76.61 \\
		\hline
		TraceNorm & 0.6987 & 0.6069 & 0.0485 & 521.56 \\
		\hline
		MaxNorm & 0.6615 & 0.6646 & 0.1303 & 71.49 \\
		\hline
		EB1 & 0.6679 & 0.6061 & 0.0273 & 138.15 \\
		\hline
		EB2 & 0.6770 & 0.5969 & 0.0344 & 145.59 \\
		\hline
	\end{tabular}
\end{table}

\begin{figure}[tbp]
	\centering
	\begin{tikzpicture}
		\begin{axis}[
			width=0.65\linewidth,height=0.65\linewidth,
			xmin=0,xmax=1,ymin=0,ymax=1,
			xtick={0,0.2,0.4,0.6,0.8,1},ytick={0,0.2,0.4,0.6,0.8,1},
			tick label style={font=\scriptsize},
			label style={font=\small},title style={font=\small},
			xlabel={Mean predicted probability},ylabel={Observed proportion of ones},
			grid=major,grid style={gray!20},
			legend columns=5,
			legend to name=movieupdatedlegend,
			]
			\addplot[black,dashed,forget plot] coordinates {(0,0) (1,1)};
			\addplot[color=blue,mark=o,thick,mark size=2pt] coordinates {
				(0.02980083,0.18282548)
				(0.15313388,0.25676613)
				(0.25406161,0.29622266)
				(0.35579279,0.34794908)
				(0.46009767,0.44451395)
				(0.54419978,0.54306952)
				(0.64765494,0.63337266)
				(0.74736687,0.73393787)
				(0.84776112,0.79983156)
				(0.96162946,0.87935977)
			};
			\addlegendentry{MMGN}
			\addplot[color=black,mark=square,thick,mark size=2pt] coordinates {
				(0.04863288,0.15637134)
				(0.15153481,0.22486170)
				(0.25114504,0.28926523)
				(0.35156671,0.36473963)
				(0.45108570,0.44509410)
				(0.54965761,0.55073490)
				(0.64980470,0.63497674)
				(0.75060498,0.69576149)
				(0.85060212,0.75698179)
				(0.95736327,0.86378036)
			};
			\addlegendentry{TraceNorm}
			\addplot[color=green!60!black,mark=triangle,thick,mark size=2pt] coordinates {
				(0.16184885,0.36363636)
				(0.25844409,0.24444444)
				(0.36928199,0.20769231)
				(0.48055565,0.37392483)
				(0.52744879,0.66527350)
				(0.63193852,0.86466848)
				(0.73255332,0.90688259)
				(0.83055179,0.87500000)
				(0.91470417,1.00000000)
			};
			\addlegendentry{MaxNorm}
			\addplot[color=red,mark=*,very thick,mark size=2pt] coordinates {
				(0.07401970,0.07643312)
				(0.16075518,0.15481172)
				(0.25686839,0.21604938)
				(0.35591570,0.31481481)
				(0.45434420,0.43396598)
				(0.54865358,0.55797041)
				(0.64694032,0.69448229)
				(0.74571547,0.78132765)
				(0.84389338,0.86745073)
				(0.93041359,0.94936709)
			};
			\addlegendentry{EB1}
			\addplot[color=magenta,mark=diamond*,very thick,dashed,mark size=2pt] coordinates {
				(0.08007304,0.06802721)
				(0.16139645,0.10075330)
				(0.25676226,0.20232733)
				(0.35575023,0.29748241)
				(0.45408101,0.44158809)
				(0.54896039,0.56768821)
				(0.64662247,0.69667543)
				(0.74608476,0.79229391)
				(0.84468918,0.88168057)
				(0.93276545,0.95182013)
			};
			\addlegendentry{EB2}
		\end{axis}
	\end{tikzpicture}
	\par\vspace{2mm}
	\pgfplotslegendfromname{movieupdatedlegend}
	\caption{Reliability diagrams on the MovieLens 100K dataset.
		Each point shows the mean predicted probability and the observed proportion of ones
		within a nonempty probability bin, evaluated on held-out observed entries.
		The dashed line indicates perfect calibration.}
	\label{fig_movie}
\end{figure}

\section{Conclusion}\label{sec_concl}
In this study, we developed empirical Bayes (EB) algorithms for 1-bit matrix completion, motivated from the singular value shrinkage estimator for a normal mean matrix by \cite{Efron72}.
Simulation studies and real-data applications demonstrate that the proposed method achieves competitive predictive accuracy and favorable predictive calibration.

There are several potential directions for future research.
First, we assumed that the observed entries $\Omega$ are selected independently of the latent variables, which corresponds to the Missing Completely at Random (MCAR) mechanism. 
In many practical scenarios, however, the missingness may depend on the latent variables, categorized as Missing at Random (MAR) or Missing Not at Random (MNAR). 
Incorporating explicit models for the missing mechanism into the current framework would be effective for several applications such as causal inference. 
Second, while we employed the Monte Carlo EM in this study, it is essential to develop a computationally more efficient implementation of the EB algorithm for large-scale applications, such as variational Bayes techniques \citep{Hui}. 
Another direction is to incorporate regularized estimation of the prior covariance into the proposed empirical Bayes framework, which may improve performance when $q$ is comparable to $p$.
Finally, extending the singular value shrinkage framework to higher-order tensors or multi-view 1-bit data presents another interesting direction for future work.

\section*{Acknowledgement}
We thank the referees for constructive comments.
We thank Kaito Fujii, Francis Hui and Shonosuke Sugasawa for helpful comments.
We used LLMs for improving the clarity of expression during manuscript preparation.
Takeru Matsuda was supported by JSPS KAKENHI Grant Numbers 22K17865, 24K02951, 25K22750, and 26H02517, and JST Grant Numbers JPMJMS2024 and JPMJAP25B1.

\appendix

\section{Derivation of the M-step}
Recall that, for parameter estimation of a statistical model $p(X,Z \mid \theta)$ with observed variable $X$ and unobserved variable $Z$, the EM algorithm iterates the update
\begin{align}
	\theta_{t+1} = \argmax_{\theta} Q(\theta,\theta_t) \label{EM_update}
\end{align}
until convergence, where
\begin{align}
	Q(\theta,\theta_t) = {\rm E}_{Z \mid X,\theta_t} [\log p(X,Z \mid \theta)]. \label{EM_Q}
\end{align}
In Monte Carlo EM, the expectation in $Q(\theta,\theta_t)$ is approximated by Monte Carlo methods.
In the current setting, $X$, $Z$, and $\theta$ correspond to $Y_\Omega$, $M$, and $\Sigma$, respectively. 

From
\begin{align*}
	\log p(Y_{\Omega}, M \mid {\Sigma}) &= \log p(M \mid \Sigma) + \log p(Y_{\Omega} \mid M) \\
	&= -\frac{pq}{2} \log (2\pi) -\frac{p}{2} \log \det {\Sigma} - \frac{1}{2} \sum_{i=1}^p m_i^{\top} {\Sigma}^{-1} m_i \\
	& \qquad + \sum_{(i,j) \in \Omega} \left( Y_{ij} \log f(M_{ij}) + (1-Y_{ij}) \log (1-f(M_{ij})) \right),
\end{align*}
the Q-function \eqref{EM_Q} is given by
\begin{align*}
	Q({\Sigma},\Sigma_t) &= {\rm E}_{M \mid Y_{\Omega},\Sigma_t} [\log p(Y_{\Omega}, M \mid {\Sigma})] \\
	&= -\frac{pq}{2} \log (2\pi) -\frac{p}{2} \log \det {\Sigma} - \frac{1}{2} {\rm tr} \left( {\Sigma}^{-1} \cdot {\rm E}_{M \mid Y_{\Omega},\Sigma_t} \left[ \sum_{i=1}^p m_i m_i^{\top} \right] \right) \\
	& \qquad + {\rm E}_{M \mid Y_{\Omega},\Sigma_t} \left[ \sum_{(i,j) \in \Omega} \left( Y_{ij} \log f(M_{ij}) + (1-Y_{ij}) \log (1-f(M_{ij})) \right) \right]. 
\end{align*}
By maximizing it with respect to $\Sigma$, the parameter update \eqref{EM_update} is given by \eqref{Mstep}.

\end{document}